\pdfoutput=1
\documentclass{article}

\usepackage{microtype}
\usepackage{graphicx}
\usepackage{subfigure}
\usepackage{booktabs} 
\usepackage[table]{xcolor}
\usepackage{amsmath}
\usepackage{amsfonts}
\usepackage{hyperref}
\usepackage{xcolor}
\usepackage{float}
\usepackage{multirow}
\usepackage{threeparttable}

\usepackage[accepted]{mlsys2025}

\mlsystitlerunning{\name: Efficient Quantized MoE Inference with Mixture of Low-Rank Compensators}

\usepackage{xspace}

\NewDocumentCommand{\var}{O{s} m O{}}{%
  \ensuremath{#1_{#2}^{#3}}
}
\usepackage{siunitx}

\newcommand{\commentout}[1]{}

\definecolor{light-gray}{gray}{0.80}

\newcommand\fref{Fig.~\ref}
\newcommand\tref{Table~\ref}
\newcommand\sref{\S~\ref}

\def\V{{\mathbb{V}}}

\usepackage{amsthm}

\usepackage{amsthm}

\newtheorem{mytheorem}{Theorem}
\newtheorem{assumption}[mytheorem]{Assumption}
\newtheorem{lemma}[mytheorem]{Lemma}
\newtheorem{remk}[mytheorem]{Remark}

\newcommand{\name}{MiLo\xspace}
\def\W{{\mathbf W}}
\def\U{{\mathbf U}}
\def\V{{\mathbf V}}
\def\E{{\mathbf E}}

\usepackage{tikz}
\usetikzlibrary{shapes.misc}

\usepackage[skip=2pt]{caption}
\usepackage{setspace}
\usepackage{enumitem}
\usepackage{titlesec}

\setlist{nosep} 
\newlength{\subsectionbelowskip}
\newlength{\subsectionaboveskip}

\newlength{\paragraphaboveskip}

\newcommand{\setvspace}[2]{%
  #1 = #2
  \advance #1 by -1\parskip}

\titlespacing*{\subsection}{0pt}{\subsectionaboveskip}{\subsectionbelowskip}
\titlespacing*{\subsubsection}{0pt}{\subsectionaboveskip}{\subsectionbelowskip}
\titlespacing*{\paragraph}{0pt}{\paragraphaboveskip}{*}

\makeatletter 
\def\thm@space@setup{%
  \thm@preskip=3pt
  \thm@postskip=\thm@preskip 
}
\makeatother

\setlist[itemize]{noitemsep, topsep=0pt}

\makeatletter
\g@addto@macro\normalsize{%
  \setlength\abovedisplayskip{1pt}
  \setlength\belowdisplayskip{1pt}
  \setlength\abovedisplayshortskip{1pt}
  \setlength\belowdisplayshortskip{1pt}
}
\makeatother

\makeatletter 
\renewenvironment{proof}[1][\proofname]{\par
  \vspace{1pt}
  \pushQED{\qed}%
  \normalfont
  \topsep0pt \partopsep0pt 
  \trivlist
  \item[\hskip\labelsep
        \itshape
    #1\@addpunct{.}]\ignorespaces
}{%
  \popQED\endtrivlist\@endpefalse
  \addvspace{3pt plus 3pt} 
}
\makeatother 

\begin{document}

\twocolumn[
\mlsystitle{\name: Efficient Quantized MoE Inference with Mixture of Low-Rank Compensators}



\mlsyssetsymbol{equal}{*}

\begin{mlsysauthorlist}
\mlsysauthor{Beichen Huang}{equal,uiuc,intern}
\mlsysauthor{Yueming Yuan}{equal,uiuc}
\mlsysauthor{Zelei Shao}{equal,uiuc}
\mlsysauthor{Minjia Zhang}{uiuc}
\end{mlsysauthorlist}

\mlsysaffiliation{intern}{Work done while intern at UIUC}
\mlsysaffiliation{uiuc}{SSAIL Lab, Department of Computer Science, University of Illinois Urbana-Champaign, Urbana, United States}
\mlsyscorrespondingauthor{Minjia Zhang}{minjiaz@illinois.edu}

\mlsyskeywords{Quantization, Low Rank Matrix, Mixture of Experts, Kernel, CUDA, MLSys}

\vskip 0.3in

\begin{abstract}
A critical approach for efficiently deploying Mixture-of-Experts (MoE) models with massive parameters is quantization. However, state-of-the-art MoE models suffer from non-negligible accuracy loss with extreme quantization, such as under 4 bits. To address this, we introduce \name, a novel method that augments highly quantized MoEs with a mixture of low-rank compensators. These compensators consume only a small amount of additional memory but significantly recover accuracy loss from extreme quantization. \name also identifies that MoE models exhibit distinctive characteristics across weights due to their hybrid dense-sparse architectures, and employs adaptive rank selection policies along with iterative optimizations to close the accuracy gap. \name does not rely on calibration data, allowing it to generalize to different MoE models and datasets without overfitting to a calibration set. To avoid the hardware inefficiencies of extreme quantization, such as 3-bit, \name develops Tensor Core-friendly 3-bit kernels, enabling measured latency speedups on 3-bit quantized MoE models. Our evaluation shows that \name outperforms existing methods on SoTA MoE models across various tasks. The MiLo code is open-sourced on GitHub: \url{https://github.com/Supercomputing-System-AI-Lab/MiLo}.
\end{abstract}
]



\printAffiliationsAndNotice{\mlsysEqualContribution} 

\section{Introduction}
\label{sec:intro}

Large Language Models (LLMs) have demonstrated remarkable success across various natural language processing tasks, including language understanding, reasoning, and generation \cite{gpt-3,gpt-4,gpt4o,openai-o1}. However, further scaling the models poses significant challenges to computational resources and memory consumption~\cite{scaling-law-nlp,megatron-lm-v2,gemini}. Mixture-of-Experts (MoE) has emerged as a promising solution. By incorporating sparsely activated expert layers, MoE allows scaling up LLM parameters while maintaining a similar compute requirement~\cite{switch-transformer,glam,meta-moe,deepspeed-moe,dai2024deepseekmoe,jiang2024mixtral}. 

Despite its promising results, MoE models face severe memory challenges that hinder practical deployment.  For example, the Mixtral-8$\times$7B MoE model \cite{jiang2024mixtral} requires $\sim$90GB of memory to just host the model weights in half-precision, while an NVIDIA A100 only has 40/80GB memory. More recent MoEs, such as the Arctic MoE~\cite{snowflake-moe}, further push the MoE boundaries with their massive scale. With a staggering 480B parameters, these MoEs require an immense amount of memory, e.g., close to 1TB, to deploy effectively. 

When the memory usage exceeds GPU capacity, the inference of MoEs can resort to offloading~\cite{eliseev2023fast} or multi-GPU inference~\cite{deepspeed-inference}. While these methods help mitigate the pressure on the scarce GPU memory from hosting MoE models, offloading to CPU/NMVe adds non-trivial overhead to inference latency due to limited PCIe bandwidth, and multi-GPU inference significantly increases the hardware cost of deploying MoEs. 

Among different approaches, model quantization techniques have been demonstrated as a promising technique to compress LLMs~\cite{frantar2022gptq,autogptq,smoothquant,lin2024awq}. However, applying existing quantization methods to MoE models is hard: 

\begin{figure*}
    \centering
    \includegraphics[width=1\linewidth]{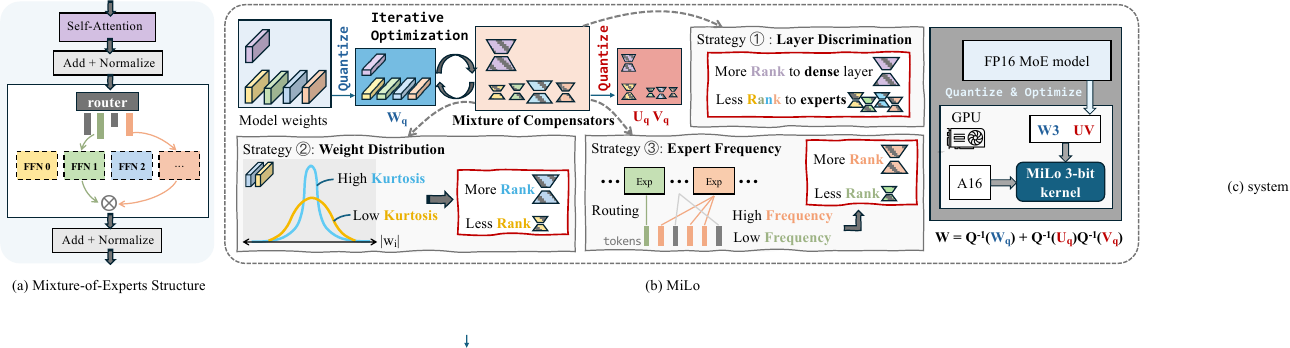}
    \vspace{-5pt}
    \caption{Overview of our \name approach. At a high level, \name employs a \emph{Quantize-Then-Compensate} approach, which augments low-bit quantized MoEs with a mixture of low-rank compensators, whose ranks are adaptively decided based on the distinctive characteristics of MoE weights. To minimize the accuracy loss, \name introduces an iterative optimization algorithm that jointly optimizes quantized MoEs and the mixture of low-rank compensators. \name includes a set of hardware-friendly INT3 kernels to achieve high measured speedups.}
    \label{fig:overview}
\end{figure*}

\begin{itemize}
    \item \textbf{SoTA MoE models suffer from non-negligible accuracy loss with extreme quantization, e.g., under 4 bits.} Traditional quantization-aware training is hard to apply to LLMs due to its high training cost~\cite{zeroquant,dettmers2022gpt3}. Recent post-training quantization works, such as GPTQ~\cite{frantar2022gptq,autogptq} and AWQ~\cite{lin2024awq} have demonstrated their effectiveness for dense LLMs towards 4-bit compression. However, further pushing the quantization limit to under 4-bit, e.g., 3-bit, leads to major performance loss. Tab.~\ref{tab:AWQ_GPTQ} reports the INT4 quantization time and the Wikitext2 perplexity results from GPTQ and Round-To-Nearest (RTN). INT4 quantization generally leads to a minor loss of accuracy in all the settings, but directly applying INT3 quantization to MoE model cannot lead to a satisfying accuracy. 
\begin{table}[htbp]
\small
  \centering
  \caption{Comparison of existing quantization methods.}
    \begin{tabular}{ccccc}
    \toprule
    Wikitext2-PPL$\downarrow$ & Quant-time & FP16  & INT4  & INT3 \\
    \midrule
          & \multicolumn{4}{c}{Mixtral-8$\times$7B} \\
\cmidrule{2-5}    RTN   & 321s      & 3.42      &   3.63    & 4.81 \\
    GPTQ  & 5315s     & 3.42  & 3.63  & 4.61 \\
    \midrule
          & \multicolumn{4}{c}{DeepSeek-MoE} \\
\cmidrule{2-5}    RTN   & 91s      & 5.83      & 6.04      &7.32  \\
    GPTQ  & 3355s      & 5.83  & 6.02  & 7.08 \\
    \bottomrule
    \end{tabular}%
  \label{tab:AWQ_GPTQ}%
\end{table}%

    \item \textbf{State-of-the-art methods suffer from calibration data bias and prolonged quantization time, making them hard to apply to MoEs with massive parameters.} Prevailing quantization methods, such as GPTQ and AWQ, rely on calibration data to obtain high accuracy. However, the choice of the calibration data introduces a bias, which causes overfitting during quantization. This is less desirable because recent MoE-based LLMs are still generalist models. Furthermore, calibration requires forward propagation to gain information from input dataset, which is computationally intensive and time-consuming, making it difficult to test on MoE models with massive parameters. 
    \item \textbf{Difficulty of converting theoretical savings from extreme quantization to measured speedups for MoEs, especially with INT3 weight-only quantization and batch size $>$1.} While recent work reported the accuracy of 3-bit quantized MoE~\cite{eliseev2023fast,li2024examining}, most of these work do not report latency improvement. To be specific, existing work often does not discuss the weight packing and de-quantization cost associated with 3-bit quantization scheme, which in fact has a big impact on the performance benefit of using 3-bit quantization.
\end{itemize}

These challenges signify the need for a more advanced optimization method for MoE models. We start from some intriguing observations (\sref{sec:observations}) that MoE models exhibit distinct characteristics across different weights. In particular, we observe that there are distinct patterns among parameters in non-expert layers and sparsely activated experts, as well as across different experts. Additionally, while INT3 quantization effectively captures outliers, information loss tends to occur at relatively insignificant weight values, motivating error reconstruction methods.

Based on this observation, we propose \name to \emph{compress MoEs by augmenting low-bit quantized MoEs with a \underline{Mi}xture of \underline{Lo}w-rank compensators}. First, to avoid overfitting and the expensive calibration overhead, we employ a calibration-free quantization algorithm to obtain extreme quantized MoEs, e.g., INT3 MoE. Second, we compensate low-bit quantized MoEs with a mixture of decomposed residual matrices, i.e., the mixture of low-rank compensators, to recover the information loss with a tiny portion of memory overhead. We show that such a mixture of low-rank compensators is quite powerful, which enables an adaptive rank selection strategy based on model structures and data distributions, and can be quantized to further reduce their memory consumption without hurting effectiveness. Thirdly, we enhance quantization performance with an iterative optimization algorithm that jointly optimizes quantized MoEs and their compensators. Finally, we develop hardware-friendly Mixed Precision 3-bit GeMM kernel, using zero-bit-waste INT3 weight packing, binary manipulation based de-quantization, and multi-level pipelining to achieve high measured speedups. Our contributions are: 

\begin{itemize}
    \item We propose \name, a novel algorithm that effectively compresses MoE models with INT3 quantization and mixture of adaptive low-rank compensators. \name is training-free and does not suffer from calibration bias.
    \item We propose an efficient INT3$\times$ FP16 Mixed Precision GeMM CUDA kernel,  for the first time, we demonstrate that it is possible to allow SoTA MoEs to achieve measured latency 1.2x speedups than SoTA backend MARLIN with batch size $>$ 1.      
    \item We evaluate \name on SoTA MoE models Mixtral-8$\times$7B and DeepSeek-MoE, and our evaluation results show that \name effectively compresses MoE models with negligible accuracy loss, i.e., recovering over 87\% of accuracy on Wikitext2 perplexity with 22\% compression ratio. Notably, \name achieves up to 3$\times$ speedups compared with baseline approaches.  
\end{itemize}

\section{Related Works}
\label{sec:related}
 
\textbf{Post-Training Quantization (PTQ) for LLMs.} In a broad taxonomy, there are two setting of PTQ: quantizing both weight and activation~\cite{dettmers2022gpt3,yao2022zeroquant,smoothquant} and weight-only quantization~\cite{park2022lut,dettmers2023case}. 
We focus on the second setting in this work, as weights are the primary memory bottleneck for MoEs. In this line of work, the state-of-the-art methods, such as GPTQ \cite{frantar2022gptq,autogptq} and AWQ~\cite{lin2024awq}, manage to compress dense LLMs to 4-bit without losing much accuracy. However, existing methods often resort to calibration data to minimize the error in layer outputs caused by outliers. Calibration-based methods suffer from over-fitting to the calibration set and long quantization time. More recently, researchers have also explored calibration-free PTQ methods, such as HQQ \cite{badri2023hqq}. HQQ captures outliers using a hyper-Laplacian distribution with closed-form solutions. However, we show that existing calibration-free PTQ methods fall short in capturing insignificant weight values.   

\textbf{Low-rank methods for LLM compression.} Low-rank factorization techniques, i.e. SVD, are applicable to many aspects of LLM compression. ASVD \cite{yuan2023asvd} uses activation to identify the salient weight, and decomposes the rest weight to low-rank matrices to compress the model. GFM \cite{yu2023compressing} aims at decomposing the features. A recent work named LoRC \cite{yao2024exploring} brings the low-rank factorization method to the error matrix between the vanilla weight and quantized weight, and treats it as compensation to the quantization. Different from those efforts, we explore a mixture of low-rank compensators for MoE models by considering their unique characteristics.  

\textbf{MoE compression.} Early work on MoE quantization focuses on translation tasks~\cite{kim2023mixture}. Some heuristic strategies of mix-precision quantization for MoE are investigated in \cite{li2024examining}, revealing the bit-sensitivity of different MoE blocks. One extreme example is \cite{frantar2024qmoe}, which aims at a sub-1-bit compression through algorithm and compression format co-design. However, multiple studies show that even 3-bit quantization hurts MoE model accuracy significantly~\cite{eliseev2023fast,li2024examining}. 
On a separate line of research, researchers have also investigated pruning experts~\cite{chen2022task,li2023merge}, which is complementary to MoE quantization.

\textbf{System support for low-bit quantized LLMs.} TensorRT-LLM has the SoTA kernel support for weight-only quantization. However, it only supports weights in INT4 (W4A16) or INT8 (W8A16 and W8A8)~\cite{tensorrt}.
In contrast, we provide additional support for W3A16. Additionally, Bitsandbytes supports W8A8~\cite{bitsandbytes}. AWQ has GPU kernel implementation for W4A16~\cite{lin2024awq}.
Llama.cpp supports 2/3/4/5/6/8-bit quantization on CPUs and GPU~\cite{llama-cpp}. However, their kernels cannot make effective use of Tensor Cores. More recently, MARLIN~\cite{marlin} kernels have been developed to support W4A16 quantized GeMM calculation by maximizing the usage of different hardware units on NVIDIA GPUs.
GPTQ has a basic GeMV W3A16 implementation for batch size 1 (e.g., memory-bound scenarios) using vector intrinsics, e.g., \texttt{\_\_hfma2}.
But it does not support batch size $>$ 1. To the best of our knowledge, this work is the first that supports W3A16 on Tensor Core with batch size $>$ 1 with measured speedups on MoE models.

\section{Methodology}

\subsection{Rationale}
We propose the method motivated by the layer-divergence nature of sparse models and the observation of the low-bit quantization degradation in Mixtral-8$\times$7B~\cite{jiang2024mixtral} and DeepSeek-MoE~\cite{dai2024deepseekmoe}.

\subsubsection{Observations}
\label{sec:observations}

\begin{figure}[t]
    \centering
    \includegraphics[width=1\linewidth]{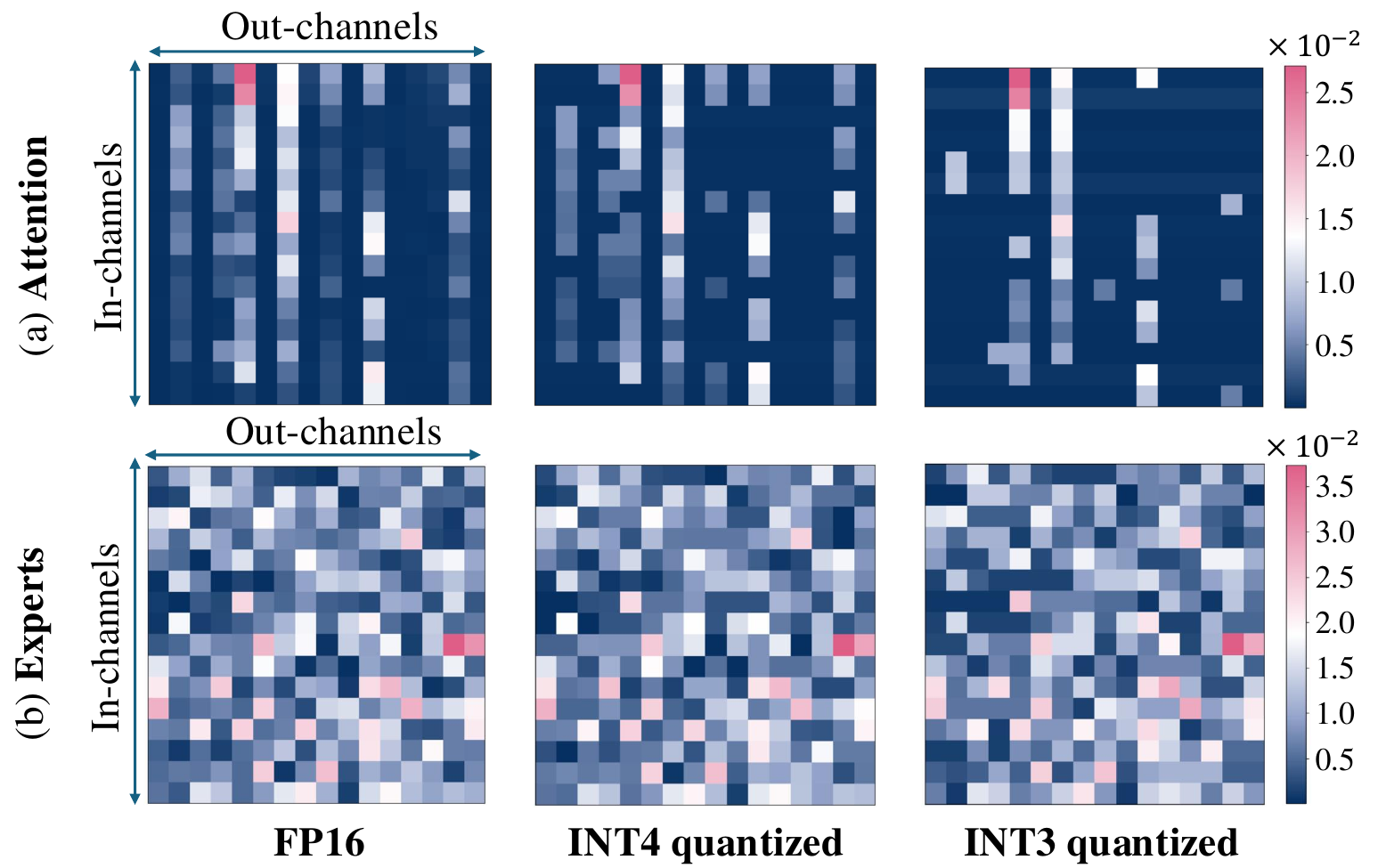}
    \caption{Mixtral-8x7B's (a) weight sampling from \emph{attention} projection and (b) weight sampling from \emph{expert}.}
    \label{fig:heatmap}
\end{figure}

\textbf{Observation 1: Parameter divergence of sparse models.}
We observe that MoE models exhibit varying characteristics across weights. Since the weights are trained on different amounts of data, the properties of each layer may diverge within a transformer model. For example, \fref{fig:heatmap} illustrates distinct patterns between the parameters in the attention projection and the expert weights.

(1) \emph{Dense layers differ from sparsely activated ones.} During training, the dense layers and sparse expert layers were fed with different 
 amounts of tokens. Dense layers include the attention projections and the dense components (e.g., shared experts) in hybrid architectures. \fref{fig:heatmap} and \fref{fig:distribution-attention} show the case in Mixtral-8$\times$7B~\cite{jiang2024mixtral}. The attention weight distributions are more heavy-tailed with outliers along the channel-wise dimension.

This property can be also captured by the \emph{Kurtosis} of the matrix, defined as $K = \frac{\mathbb{E}[(X - \mu)^4]}{\sigma^4}$. Higher Kurtosis indicates a more heavy-tailed distribution, which reflects the number of outliers in a matrix~\cite{li2024evaluatingquantizedlargelanguage}. Tab.~\ref{table:layer-characteristics} shows the average Kurtosis across weights for different blocks, revealing dense structures have more tail values than sparse layers. 

\begin{table}[H]
\caption{Kurtosis and average residual matrix rank across layers and models. The rank is measured by the number of singular values $\sigma_i$ smaller than $\tau\cdot\sigma_{\max}$, where we use $\tau = 0.5$ in this table. A: Attention projection weights, E: sparse expert weights, SE: shared expert weights in DeepSeek. D represents \emph{densely} activated layers, S represents \emph{sparsely} activated layers.}
\label{table:layer-characteristics}
\vskip 0.1in
\begin{center}
\newcommand{\smallcolspc}{\hspace*{0.28em}}
\begin{small}
\begin{sc}
\begin{tabular}{@{\smallcolspc}c@{\smallcolspc}|c|c|c|c|c}
\toprule
 & \multicolumn{2}{c|}{Mixtral-8$\times$7B} & \multicolumn{3}{c}{DeepSeek-MoE} \\
layer & A(D) & E(S) & A(D) & SE(D) & E(S)\\
\midrule
Kurtosis & 1.57 & -0.53 & 0.016 & 0.32 & -0.89 \\
\midrule
Res. Rank & 514 & 1730 & 438 & 286 & 602 \\
\bottomrule
\end{tabular}
\end{sc}
\end{small}
\end{center}
\vskip -0.15in
\end{table}

\begin{figure}[t]
    \centering
    \includegraphics[width=1\linewidth]{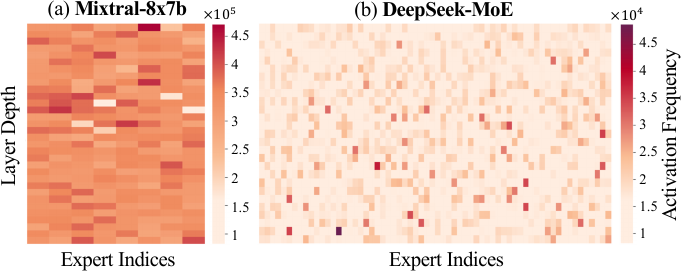}
    \caption{Heatmap of expert activation frequency in Mixtral-8$\times$7B and DeepSeek-MoE on the WikiText-2 task. The vertical axis from top to bottom represents the layer depth, and the horizontal axis represents expert indices.}
    \label{fig:expert-frequency}
\end{figure}

(2) \emph{Not all the experts are equal.} Within an MoE layer, the experts are also trained on different subsets of tokens, which cause characterization divergence among experts. Besides, the experts in an MoE layer are not equally activated at identical frequency at all times. As plotted in \fref{fig:expert-frequency}, the expert frequency diverges, especially for fine-grained settings. In DeepSeek-MoE, the most frequently activated expert is activated $11.7 \times$ more often than the least activated expert within the same layer.

\textbf{Insight.} The diverse patterns in MoE pose unique challenges and opportunities for MoE compression, motivating novel approaches to utilize them effectively. 

\textbf{Observation 2: Low-bit quantization's degradation in insignificant weight values.}
\fref{fig:heatmap} shows a sampling of un-quantized half-precision weights and de-quantized INT3 weights from an expert layer and a self-attention layer in Mixtral-8$\times$7B. Interestingly, the INT3 quantization captures the extreme values, and information loss mainly occurs at relatively \emph{insignificant weight values}. In other words, quantizations capture the outliers adequately while sacrificing the representation of the moderate values as a tradeoff. 

Notably, the layer-divergence also plays a role in this effect. The layers with a high Kurtosis, such as the Attention layer in Mixtral-8$\times$7B, suffer more from low-bit quantization due to their heavy-tailed nature. This effect can be observed more in \fref{fig:heatmap}, where the INT3 quantized weight of attention projection (top right) shows a greater loss of information compared to the expert projection. 
The residual matrix is an important indicator for analyzing the quantization error. In Table \ref{table:layer-characteristics}, we measure the residual matrix rank (the number of singular values $\sigma_i$ smaller than $\tau\cdot\sigma_{\max}$) across layers in Mixtral-8$\times$7B and DeepSeek-MoE, where the rank demonstrates negative correlation to the Kurtosis.

\textbf{Insight.} Extreme quantization is able to capture the outliers in MoE weights at the sacrifice of the expressiveness of insignificant weight values. We need to come up with a method to recover the information loss of those values, with the objective of fully recovering the original FP16 model quality. Ideally, the method should only add slightly more memory while efficiently representing the lost information in the extreme quantization scenario.

\subsubsection{Low-rank Error Construction}

\begin{figure*}[!ht]
    \centering
    \includegraphics[width=\textwidth]{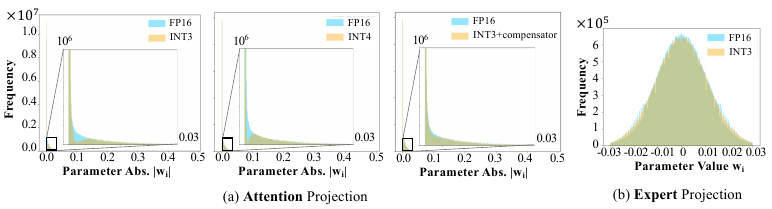}
    \caption{The overlapping region of quantized and half-precision distribution in each figure is shown in green. (a) Information loss analysis for attention layer Mixtral-8$\times$7B. \textbf{Left}: INT3 weight quantization captures the outliers adequately but has large information loss at relatively insignificant weight values. \textbf{Middle}: INT4 is able to close some of the information gap but not completely. \textbf{Right}: INT3 together with low-rank matrices manage to close the information loss gap. (b) Information loss for expert layer at same $|w_i|$ range. 
    }
    \label{fig:distribution-attention}
\end{figure*}

In this work, we consider the solution \emph{residual reconstruction}, which represents the missing information aside from the quantized matrices and corrects quantization errors. 
Based on the preceding analysis, to complement the quantization that captures the outlier information, we expect a method to estimate the residual that captures \emph{the moderate values} in a matrix well. Also, we require the method to work together with the quantization - they should be optimized together and avoid representation redundancy. 

Based on the observations and analysis, we consider \emph{low-rank compensation(LoRC)}~\cite{lorc} as a countermeasure. Low-rank compensation reconstructs the residual of quantization using a low-rank estimation. The weight after compensation is $\Tilde{\mathbf{W}}_{LoRC} = Q^{-1}({\W}_q) + \U\V$, where $\W_q$ is the quantized weight, and $\U \in \mathbb{R}^{m \times r}$, $\V\in \mathbb{R}^{r \times n}$ is optimized to let $\U\V$ closely approximate the residual $\E = \W - \W_q$. 
This optimization utilizes SVD on the residual matrix, where $\E = \U\Sigma \V$, then obtain $\U, \V$ by keeping the largest $r$ singular values in $\Sigma$. \fref{fig:distribution-attention}(c) shows that INT3 plus low rank compensation is able to recover most information loss.

However, directly applying low-rank to an optimized quantization is a suboptimal solution. Since we expect low-rank to represent part of the information, the quantization itself should also be optimized to adapt to the ``low-rank residual". Also, specific considerations based on the sparse nature of MoE are required, or the uniform low-rank compensation would introduce a large memory consumption. To solve these problems, we propose the \name\ method.

\subsection{\name}

In previous sections, we analyze the quantization and compensation separately, highlighting the their potential and challenges for compressing MoE models while establishing the building block for the proposed method. In this section, we bring two parts together. 

Formally, the goal is to design an algorithm that minimizes the performance drop of MoE models after compression (without fine-tuning) with respect to their uncompressed counterparts. 
For a pre-trained MoE model $f(x;W)$, we propose to solve the following optimization problem $\mathcal{P}$:
\begin{equation}
\label{eq:orig}
    \mathop{\arg\min}\limits_{z,s,U,V} \mathcal{L}(W - Q^{-1}_{z,s}(Q_{z,s}(W)) - UV)
\end{equation}
where $\mathcal{L}$ is loss function, $W$ is the original weight, and $U,V$ are low-rank matrices to approximate the error, i.e., low-rank compensator. $Q$ and $Q^{-1}$ are the quantization operator and de-quantization operator, defined as: 
\begin{align}
    W_q = Q_{z,s}(W) &= round((W-UV)/s+z)\label{eq:wq}
 \\ W_{dq} = Q^{-1}_{z,s}(W_q) &= s(W_q - z) \label{eq:wdq}
\end{align}
where $s$ and $z$ are vectors of scaling parameter and zero-point for the quantizer, respectively. Meanwhile, the optimization should be subject to the constraint $rank(UV) \leq r$, i.e., $UV$ has a rank that does not exceed a specified threshold $r$.

\subsubsection{Decomposing the optimization into sub-problems} 
The above problem is non-differentiable with combinatorial constraints which cannot be solved with stochastic gradient descent methods (e.g., Adam~\cite{adam-optimizer-arvix}). We present an optimization algorithm, which decomposes the problem $\mathcal{P}$ into two distinct subproblems. \emph{Subproblem 1 (sp1): quantization error minimization}, which aims to reduce the discrepancy between $W - UV$ and its quantized version $W_q$, and \emph{subproblem 2 (sp2): low-rank compensation maximization}, which focuses on finding low-rank matrices $U$ and $V$ such that $UV$ closely approximate the quantization residual matrix $W - W_{dq}$. We then alternatively solve the subproblems until convergence.

\subsubsection{Optimizing $\mathbf{W_q}$ with U,V fixed}
\label{subsubsec:hqq}

In iteration $t$ of $\mathcal{P}$, we first solve the \emph{sp1} by optimizing the de-quantized weights $W_{dq}^{t}$ with fixed low-rank matrices $U^{t-1},V^{t-1}$ from previous iteration. The \emph{sp1} is formulated by applying $l_{p<1}$ norm as the loss function $\mathcal{L}$ and solved as a Lagrange dual problem. Formally, it is described as:
\begin{equation}
\label{eq:hqq_obj}
    \mathop{\arg\min}\limits_{z^t,s^t} \|W - U^{t-1}V^{t-1} - W_{dq}^{t}\|_{p<1}
\end{equation}
, where $W_{dq}^{t}$ is defined as Equation \ref{eq:wdq}.
Note that at iteration 0, the matrices $U$ and $V$ are unknown, and are initialized to zero. This initialization serves as the starting point for the iterative optimization. 

 For simplicity, we fix the scaling parameter $s^t$ and only optimize the zero-point $z^t$, following the techniques in Half-Quadratic Quantization (HQQ)~\cite{badri2023hqq}. 
With an auxiliary variable $M^t$, the optimization problem \ref{eq:hqq_obj} is:
\begin{equation}
\label{eq:quant}
\small
    \mathop{\arg\min}\limits_{z^t,M^t} \|M^t\|_{p<1} + \frac{\beta}{2} \|M^t - (W - U^{t-1}V^{t-1} - W_{dq}^{t})\|_2^2
\end{equation}
Problem \ref{eq:quant} can be solved by further applying alternate optimization to update $M^t$ and $z^t$ separately, using Half-Quadratic solver \cite{120331} and generalized soft-thresholding operator \cite{badri2016non}. In each iteration $k$ of $sp1$, we first update $M_k^t$ as:
\begin{gather}
\label{eq:update_We}
    M^t_k \longleftarrow shrink_{l_p}\left(W - U^{t-1}V^{t-1} - W_{dq,k-1}^{t}),\beta\right)\hfill \\
    shrink_{l_p}\left(x,\beta\right) = sign\left(x\right)relu(|x| - \frac{|x|^{p-1}}{\beta})
\end{gather}
And then $z^t_k$ is updated as:
\begin{gather}
\label{eq:update_z}
    z^t_k \longleftarrow \langle W_{q,k}^t - \frac{(W - U^{t-1}V^{t-1} - M^{t}_{k}}{s} \rangle \\
    W_{q,k}^t = round((W-U^{t-1}V^{t-1})/s + z^{t-1})
\end{gather}
, where $\langle \cdot \rangle$ represents the average over the axis of the quantization grouping. We choose HQQ to minimize the quantization error due to its low quantization overhead, making it more scalable for large-scale models such as MoE. Moreover, we do not use any calibration data in this process, which avoids calibration data bias. We refer readers to HQQ \cite{badri2023hqq} for more detailed steps. 

\subsubsection{Solving $\mathbf{U,V}$ with $\mathbf{W_q}$ fixed} 
\label{subsubsec:svd}

With a fixed $W_{q}^t$, the problem $\emph{sp1}$ can be viewed as a standard low-rank approximation problem, which is written as:
\begin{equation}
\label{eq:compensate}
    \mathop{\arg\min}\limits_{U^{t},V^{t}} \mathcal{L}(E^t - U^{t}V^{t})
\end{equation}
, where $E^t$ is fixed as: $W - W_{dq}^t$. In the case of Frobenius norm, the problem is well studied and solved by truncated singular value decomposition, as proved in Eckart-Young-Mirsky Theorem \cite{eckart1936approximation}.
We first apply SVD to $E^{t}$ and then update $U^t,V^t$ with a given hyper-parameter rank $r$ as:
\begin{gather}
    E^t = \hat{U} \Sigma \hat{V}\\
    U^t = \hat{U}_{:,1,r}(\Sigma_{1:r,1:r})^{\frac{1}{2}};\quad
    V^t = (\Sigma_{1:r,1:r})^{\frac{1}{2}}\hat{V}_{1:r,:}\label{eq:UV} 
\end{gather}

\subsubsection{Stop Condition}
We alternate \sref{subsubsec:hqq} and \sref{subsubsec:svd} until it reaches a stop condition. We use Frobenius norm to measure the error $\epsilon_{t}$ after each iteration of $\mathcal{P}$. $\epsilon_{t}$ is defined as:
\begin{align}
\label{eq:epsilon}
    \epsilon_{t} = \| W - W_{dq}^{t} - U^{t}V^{t}\|_{F}
\end{align}
Since this provides an indirect measure of the optimization function, a monotonic decrease in $\epsilon_{t}$ is not guaranteed. In such case, we apply a sliding window average of the error over three iterations, denoted as $\hat{\epsilon}_{t}$, and stop the iteration if:
\begin{align}
\label{eq:stop_condition}
    \frac{\hat{\epsilon}_{t-1} - \hat{\epsilon}_{t}}{\hat{\epsilon}_{t-1}} < 1e^{-4}
\end{align}

In practice, we find that a few tens of iterations (e.g.,20) are sufficient for the optimization to reach a nearly converged output. Based on this, we propose an early-stop strategy, which terminates the algorithm at iteration 20 or stops the process if the error begins to diverge. This early-stop strategy is applied in all the experiments unless stated otherwise.

\subsubsection{Adaptive mixture of low-rank compensators}
\label{subsec:discriminative-low-rank}
Till now we have fixed the choice of the rank $r$ for each low-rank compensator. However, one may wonder whether the sparse nature of MoE architecture leads to more effective and efficient compression. 
Empirically, increasing the rank improves performance but also increases memory overhead. 
Rather than applying a uniform rank to all weights, a more effective strategy is to \emph{use higher ranks only where they are most effective}. The property of a matrix that reflects how changes in the rank affect the final performance is referred to as \textit{rank sensitivity} in the following discussion. To that end, we analyze from both MoE structure and matrix property perspectives for DeepSeek-MoE and Mixtral-8$\times$7B, and discuss the rank sensitivity of weights, considering memory constraints. The detailed experiment is provided in \sref{subsec:eval_analysis}.

\textbf{Rank vs. model structures.} 
In MoE models, the sparse-activation mechanism brings dense layers and sparse layers with different characteristics. Since the dense layers are always activated for input tokens, they play a more important role in the model performance. For example, the attention projections are much more rank sensitive than the linear projection with an expert. Therefore, we give \emph{dense layers} more ranks than the sparsely activated layers.

The importance of each expert also varies according to activation frequency, as experts activated more frequently contribute more significantly to specific tasks. From \fref{fig:expert-frequency}, we noticed an uneven distribution of expert activation frequencies on Wikitext2 input, particularly among DeepSeek-MoE's fine-grained experts. Therefore, expert frequency is also a good guideline for rank sensitivity.

\begin{figure}
    \centering
    \includegraphics[width=1\linewidth]{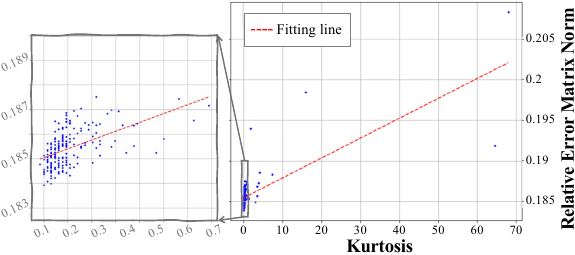}
    \caption{The correlation between relative Frobenius norm vs. Kurtosis. Each dot represents a weight matrix in layer 1 of DeepSeek-MoE.}
    \label{fig:kurtosis-vs-error}
\end{figure}

\textbf{Rank vs. data distribution.} 
As analysis in \sref{sec:observations}, we noticed that the Kurtosis reflects the outlier distribution, and the heavy-tail distributed weights suffer more in information loss under extreme quantization. Fig.~\ref{fig:kurtosis-vs-error} demonstrates the positive correlation between Kurtosis and relative quantization error $||W-W_{dq}||_F/||W||_F$. We further look into how the low-rank compensator helps in bridging the quantization error. We plot the data distribution of a self-attention layer under INT3, INT4 and INT3+LoRC in Fig.~\ref{fig:distribution-attention}(a). Compared with INT3 and INT4 quantization (in left and middle figure), the introduction of low-rank matrices refills the non-outliers, effectively compensating on a heavy-tail distributed weight. And for a weight with lower Kurtosis, as shown in Fig.~\ref{fig:distribution-attention}(b), the pattern is not that obvious. Therefore, \emph{higher Kurtosis} indicates a higher rank to bridge the information loss brought by the loss of insignificant weight.

These findings provide the foundation for constructing an adaptive mixture of low-rank compensators. Below, we outline several low-rank compensation policies. While this study primarily evaluates these specific policies, \name can readily accommodate a wide range of other strategies. 

\begin{itemize}
    \item \textbf{Uniform-\{r\}}: We set a uniform rank $r$ for all layers, including self-attention layers and expert layers. 
    \item \textbf{Dense-\{r\}}: We only set rank to dense layers, while keeping the rank of sparse layers to 0. For Mixtral-8$\times$7B model, the dense layers are self-attention layers, and for DeepSeek-MoE, dense layers contain self-attention layers, shared-experts, and dense FFN layers.
    \item \textbf{Sparse-\{r\}}: We assign rank $r$ to sparse activated layers, i.e. experts, for both Mixtral-8$\times$7B and DeepSeek-MoE models and keep rank to 0 for other layers.
    \item  \textbf{Frequency-\{r\}}: We assign higher rank to experts with higher frequency, and control the average rank to be $r$. 
    \item \textbf{Kurtosis-\{r\}}: We set higher rank to weights with higher Kurtosis, and control the average rank to be $r$.
\end{itemize}

\textbf{Insight.} Overall, we find that dense layers are the most rank sensitive structure and therefore merit higher ranks than sparse layers. The significant benefit is attributed to the fact that dense layers are activated for every token, and thus a higher rank compensator benefits all the inputs. From data distribution perspective, kurtosis and expert frequency work well in different scenarios. For models with balanced experts, e.g. Mixtral-8$\times$7B, Kurtosis is a good indicator of rank. And for those models with unbalanced experts, e.g., DeepSeek-MoE, assigning rank according to expert frequency leads to more performance improvement.


\subsubsection{Quantized low-rank mixture compensators} 
Previous paper has found that the low rank compensation matrices can be quantized to INT8 \cite{yao2024exploring}. Following this line, we reinforce this conclusion by showing that the low rank compensation matrices can be quantized to INT3 by symmetric quantization, with minor loss of accuracy. The symmetric INT3 quantization function is:
\begin{align}
\label{eq:symm_int3}
    Q_{symm}(W) = round(\frac{7\times W}{2s})+4
\end{align}
where $s$ is the scale factor, equals to the maximum value of the quantization group. Quantizing the low rank matrices to INT3 further reduces the memory overhead brought by the compensator, while retain the accuracy benefit. More results in the evaluation section \sref{subsec:eval_analysis}.



Overall, the \name algorithm is described in Algorithm \ref{alg:QC}. When performing \name to each weight, we determine the rank $r$ according to the layer structure and data distribution as analyzed in previous section, and perform the optimization until the stop condition is satisfied. The low rank matrices $U,V$ are further quantized to INT3 using symmetric quantization. The outputs of the algorithm are the zero point $z$ and INT3 low rank matrices $U,V$.

\begin{algorithm}[tb]
   \caption{\name}
   \label{alg:QC}
\begin{algorithmic}
   \STATE {\bfseries Input:} weight $W$
   \STATE Set rank $r$ from model structure or data distribution
   \STATE Initialize $U_0 = 0 , V_0 = 0$
   \REPEAT
   \STATE // Do quantization to update $z$
    \REPEAT
    \STATE Update $M^t_k$ as Equation (\ref{eq:update_We}) 
    \STATE Update $z^t_k$ as Equation (\ref{eq:update_z})
   \UNTIL the value of $W - U^tV^t - W_{dq}$ converge
   \STATE // Do compensation to update $U^t,V^t$
   \STATE Update $U^t, V^t$ as Equation (\ref{eq:UV})
   \STATE Update the error $\epsilon_t$ as Equation (\ref{eq:epsilon})
   \UNTIL Stop condition 
   is satisfied
   \STATE Quantize $U,V$ using Equation (\ref{eq:symm_int3})
   \STATE {\bfseries Output:} zero point: $z$; low rank matrices in INT3: $U,V$
\end{algorithmic}
\end{algorithm}

\subsection{Hardware-Friendly INT3 Kernel for MoE Inference}
\label{subsec:int3-kernel}

As described in \sref{sec:related}, the state-of-the-art kernel implementation for quantized GeMM is MARLIN~\cite{marlin}, which supports W4A16.
Despite demonstrating promising results, many design choices should be reconsidered when developing efficient W3A16 GeMM kernels. One of the key difficulties lies in INT3 itself: it is not a power of 2, and modern data types typically do not support INT3 values directly. \emph{How should we realize an efficient workflow for INT3 data storage, transfer and calculation of W3A16?}

\begin{figure*}
    \centering
    \includegraphics[width=1\textwidth]{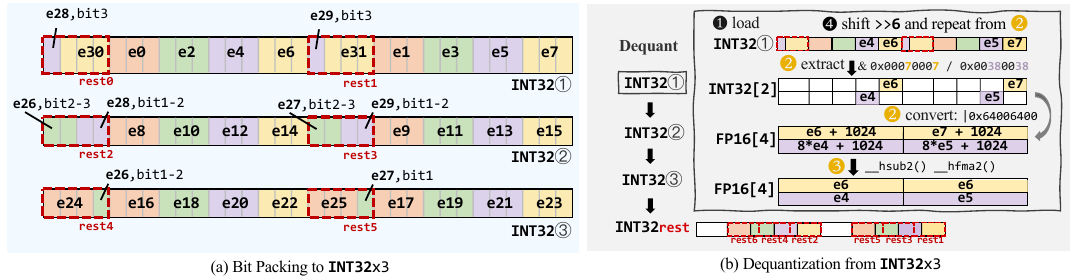}
    \caption{Figure (a) shows the zero-bit-waste 3-bit weight packing. Figure (b) shows the de-quantization process. The detailed de-quantization of INT32(1) is demonstrated. }
    \label{fig:kernel}
\end{figure*}

\paragraph{Zero bit waste 3-bit weights packing.}
We start with the INT3 weight data storage layout. 
One can pack multiple 3-bit values into a larger data type. For example, one can store ten 3-bit values in an INT32. However, this approach is inefficient as it leaves 2 bits unused. To achieve maximally efficient storage without any bit waste, we choose a packing strategy that fully utilizes each bit. The packing strategy is illustrated in \fref{fig:kernel} (a). We group every 32 consecutive INT3 weights and pack them into three INT32 values. 
In each INT32 we store 8 weights (e.g. e0, e1, \dots e7 ) and some remaining parts(e.g. rest0, rest1). By adding another 3 bit-shift operations and \texttt{|=} (i.e., bitwise OR assignment) operations, we can combine these remaining bits(represents as rest0, rest1, $\cdots$, rest5 in the figure) on the boundary into a new INT32 object that also contains 8 weights(i.e. e24, e25,\dots e31).

In addition, we perform weight reshuffling for every 16 $\times$ 64 block of weights, which corresponds to the matrix handled by a single warp to facilitate bulk loading. 
Weights are managed in groups (32 weights packed into 3 INT32s) to correctly dequantize values like e24 through e31. This requires loading data in units of 3 INT32s, which introduces an alignment issue. To address this, we split the weight matrix into two matrices: the first stores the initial two INT32s, and the second stores the last INT32.

\textbf{Efficient I2F(INT3-to-FP16) de-quantization(MiLo Dequant).}
Naively applying type-casts from INT3 to FP16 is slow. Inspired by \cite{kim-etal-2022-says}, we apply \emph{binary manipulations} to efficiently convert INT3 to FP16. Different from that work, which convert INT8/INT4 to FP16, we extend it to convert INT3 to FP16. Moreover, we convert \emph{two} INT3s to FP16s at a time, using register level parallelism, leveraging the fact that 2 FP16 elements can fit in a 32-bit register. The whole procedure for symmetric quantization is as follows, and we use INT32\textcircled{1} from \fref{fig:kernel}(b) as an example: 
\textbf{1}. Load the data into register. 
\textbf{2}. Extract [e6, e7] and [e4, e5] in another two 32-bit registers, and through binary manipulations we turn them into [1024 + e6,1024 + e7] and [1024 + 8e4,1024 + 8e5].
\textbf{3}. For symmetric quantization, we use \texttt{\_\_hsub2} and \texttt{\_\_hfma2} to get [e6-4, e7-4] and [e4-4, e5-4], while for asymmetric quantization, we would get [e6, e7],[e4, e5]
\textbf{4}. Lastly, we do the bit shift operation and repeat steps 2,3 on INT32\textcircled{2} \texttt{>>} 6 to get [e0, e1], [e2, e3]. 
In the scaling step, We use \texttt{\_\_hmul2} for symmetric quantization and \texttt{\_\_hfma2} for asymmetric quantization.

\textbf{Asynchronous global weight load.} \name leverages the asynchronous memory transfer features introduced in NVIDIA’s Ampere architecture to efficiently load neural network weights from global memory. By utilizing the \texttt{cuda::memcpy\_async} API, \name performs non-blocking transfers of weights directly into shared memory. This approach eliminates the need for threads or registers to handle the data movement, freeing them for computation. As a result, weight loading can proceed in parallel with ongoing calculations, effectively hiding the latency typically caused by accessing global memory.
  
\textbf{MoE-specific tile shape tuning.}
For certain expert layers, e.g., Mixtral-8$\times$7B have both GeMM size 4096$\times$14336 and 14336$\times$4096, the thread synchronization overhead brought by global reduction between thread blocks can be a bottleneck, and changing tile shape cuts down the number of synchronization. Therefore we enable tile shapes (256, 64), (128, 128) and (64, 256) to improve the performance.

\section{Experiment}
\label{sec:experiment}
In this section, we perform comprehensive experiments to evaluate the proposed \name and kernel. A brief implementation description is in Appendix~\ref{appendix:implementation}

\textbf{Evaluations.} The evaluation of \name is performed on 6 representative benchmarks, including language modeling(Wikitxt-2 \cite{merity2016pointer}), and common sense reasoning (PIQA\cite{bisk2020piqa}, HellaSwag\cite{zellers2019hellaswag}, Lambada \cite{radford2019language}, MMLU\cite{hendrycks2020measuring}, TriQA\cite{joshi2017triviaqa}). We report the performance on MMLU and TriQA with 5-shot and all others with zero-shot, and these results are reported in percentages. The average accuracy of zero-shot evaluation is also reported. The evaluation of the kernel of \name is performed on 3 different batch size settings.

\textbf{Baselines.} For the \name comparison, we focus on \emph{weight-only grouped} quantization because the memory consumption of MoE models is primarily dominated by the model weights, which also aligns with our motivation. All methods use a quantization group size of 64 for a fair comparison.  
\begin{itemize}
    \item RTN (round-to-nearest), which directly applies PTQ to the MoE model weights.
    \item HQQ, which is the method introduced in \cite{badri2023hqq} that uses half quadratic quantization.
    \item GPTQ, which is introduced in \cite{frantar2022gptq}. It employs Hessian information to obtain closed-form solutions for weight quantization. 
\end{itemize}

\textbf{Models.} We benchmark our method on two state-of-the-art MoEs Mixtral-8$\times$7B~\cite{jiang2024mixtral} and DeepSeek-MoE~\cite{dai2024deepseekmoe}, given that they both achieve high model quality and have severe challenges to deploy on a single GPU due to their high memory consumption. 

\begin{table*}[!ht]
  \centering
  \small
  \begin{threeparttable}
  \caption{Evaluation and Comparison of \name.}
    \label{tab:main_result}%
    \begin{tabular}{lrc|cccc|cc}
    \toprule
    W3A16 & \multicolumn{1}{l}{Memory} & Wikitext2 PPL$\downarrow$ & HellaSwag$\uparrow$ & Lambada$\uparrow$ & PIQA$\uparrow$  & Avg$\uparrow$   & MMLU$\uparrow$  & TriQA$\uparrow$ \\
    \midrule
    \hline
    \multicolumn{9}{c}{Mixtral-8$\times$7B} \\
    \midrule
    RTN   & 20.5 GB     & 4.8133  & 78.40      & 71.18      & 79.10      & 76.23      & 59.36      & 69.41 \\
    GPTQ  & \multicolumn{1}{l}{18.4 GB} & 4.7304  & 77.70  & 74.36  & 79.54  & 77.20  & 63.61  & 68.53  \\
    HQQ   & \multicolumn{1}{l}{20.5 GB} & 4.6119  & 77.88  & 69.74  & 79.16  & 75.59  & 60.93  & 70.66  \\
    \name-s1 &  20.8 GB    &   \underline{4.0335}    & \textbf{82.23}    & \underline{75.12}      & \textbf{81.33}      & \textbf{79.56}      &\underline{67.07} &\underline{75.82}\\
    \name-s2 &  21.0 GB     &   \textbf{3.9076}    &\underline{81.60}       & \textbf{75.72}      & \underline{81.12}     & \underline{79.48}      & \textbf{67.69}      & \textbf{76.42} \\
    \midrule
    \multicolumn{9}{c}{DeepSeek-MoE} \\
    \midrule
    RTN   &    {7.67 GB}   &    7.3295   &  69.81     &   65.09    &    78.29   &   71.06    &    35.03   & 50.00 \\
    GPTQ  & \multicolumn{1}{l}{6.97 GB} & 6.8234  & 73.80  & 68.62  & 77.91  & 73.44  & \text{-}\text{-}\tnote{1}     & 54.61  \\
    HQQ   & \multicolumn{1}{l}{7.67 GB} & 7.0821  & 71.38  & 66.67  & 77.25  & 71.77  & 35.63  & 54.24  \\
    \name-s1 &  7.98 GB    &   \underline{6.4226}    &  \underline{74.60}   &    \underline{71.47}   &   \underline{78.94}   &   \underline{75.00}    &    \underline{41.92}   &  \underline{59.35}\\
    \name-s2 &  8.33 GB  &  \textbf{6.2605}  &  \textbf{75.15}   &   \textbf{72.17}    &    \textbf{79.00}   &   \textbf{75.44}    &     \textbf{41.97}  &  \textbf{59.98}\\
    \bottomrule
    \end{tabular}%
    \begin{tablenotes}
    \footnotesize
    \item[1] Longer than 24hrs to run
    \end{tablenotes}
    \end{threeparttable}
\end{table*}%
\begin{table*}[!ht]
 \centering
 
  \small
  \caption{Rank Strategy Comparison under Memory Constraint.}
  \resizebox{0.9\textwidth}{!}{
  \begin{threeparttable}
    \begin{tabular}{c|ccc||ccc}
    \hline
    \multicolumn{1}{c|}{}&
    \multicolumn{3}{c||}{Model Strategy, memory constraint = 200MB} & \multicolumn{3}{c}{Sparse Layer Strategy, with Dense Layer rank = 512} \\
    \hline
    \hline
    Model  & Rank Strategy & Wikitext2 PPL$\downarrow$ & MMLU $\uparrow$ &  Rank Strategy & Wikitext2 PPL$\downarrow$ & MMLU $\uparrow$\\
   \hline
    \multirow{3}{*}{Mixtral-8$\times$7B\tnote{1}} & Uniform-28       &4.5262  &61.58   &   Uniform-32       &4.1645  &66.67    \\
                            & Dense-512         &4.1683  &65.75    &                         Kurtosis-32         & 4.1044 &67.98 \\
                            & Sparse-32         & 4.5986 &59.87 &                         Frequency-32         & 4.1698 &66.47\\
    \hline
    \multirow{3}{*}{DeepSeek-MoE\tnote{2}} & Uniform-22       & 6.9243&37.76 &   Uniform-16       & 6.4633 &40.45\\
                            & Dense-512         & 6.4743 &40.22&                         Kurtosis-16         &6.3030 &41.07 \\
                            & Sparse-24         & 6.9770 &35.93&                         Frequency-16         & 6.4570 &38.22\\
    \hline
    \end{tabular}%
    \begin{tablenotes}
    \footnotesize
    \item[1,2] Mixtral-8$\times$7B HQQ baseline: 4.6119; DeepSeek-MoE HQQ baseline: 7.0821
    \end{tablenotes}
    \end{threeparttable}
  }
  \label{tab:rank_strategy}%

 \end{table*}
\subsection{Main Results}
We propose two rank strategies with different memory consumption for both models, marked as s1 and s2, to demonstrate the effectiveness and adaptiveness of \name. The rank strategies are detailed in \tref{tab:main_rank}. 
\begin{table}[htbp]
  \centering
  \small
  \caption{Rank strategies for \name main evaluation.}
    \begin{tabular}{ccc}
    \toprule
          &       & \multicolumn{1}{c}{Rank Strategy} \\
    \midrule
    \multirow{2}[2]{*}{Mixtral-8$\times$7B} & \name-s1 & \multicolumn{1}{c}{Dense-512 + Kurtosis-16} \\
          & \name-s2 & \multicolumn{1}{c}{Dense-1024 + Kurtosis-32} \\
    \midrule
    \multirow{2}[2]{*}{DeepSeek-MoE} & \name-s1 & Dense-800 \\
          & \name-s2 & Dense-1024+Frequency-32 \\
    \bottomrule
    \end{tabular}%
  \label{tab:main_rank}%
\end{table}%

The experiment results are shown in \tref{tab:main_result}, where the best results are highlighted in bold and the second-best are underlined. All the settings achieve substantial performance gains with only a slight increase in memory usage. For Mixtral-8$\times$7B, \name-s1 improves the average zero-shot accuracy by 10\% with just 1.4\% additional memory usage compared to HQQ, while \name-s2 surpasses GPTQ by 17\% in Wikitext2 perplexity. For DeepSeek-MoE, \name-s1 delivers a direct message: a simple compensator to dense layers with a small portion of additional memory leads to huge performance improvement. And \name-s2 pushes the improvement even further, reaching 17\% of accuracy improvement in MMLU. Both the iterative algorithm and the specialized mixture of compensator strategies drive this remarkable progress, as the low-rank matrices optimization and intrinsic properties of MoE are jointly leveraged and optimized.

\subsection{Analysis Results}
\label{subsec:eval_analysis}

\begin{figure}[!ht]
    \centering
    \includegraphics[width=1\linewidth]{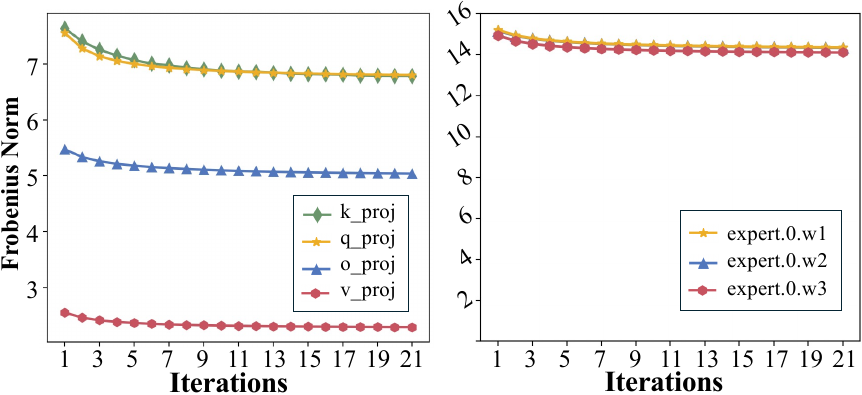}
    \caption{The change of F-norm versus iteration. (Left) The convergence curve of attention matrices. (Right) The convergence curve of expert matrices. }
    \label{fig:iter-Fnorm}
\end{figure}

\paragraph{How does the iterative optimization bring benefits?} Generally, the iterative optimization converges and leads to performance improvement. The error $\epsilon_t$, which is defined in Equation (\ref{eq:epsilon}), versus iteration is shown in Fig.~\ref{fig:iter-Fnorm}. The Frobenius norm decreases monotonically and converges at around 10 iterations. 

\paragraph{Which adaptive rank selection policy works better?} 
In \sref{subsec:discriminative-low-rank}, we discuss the policies for setting ranks in \name. Here, we compare the performance of the rank strategies as shown in \tref{tab:rank_strategy}, and a full list of evaluation results is listed in Appendix E. To focus solely on the rank strategy and eliminate iterative optimization effects, we fix the \name iterations to 1. From the model structure perspective, we compare the Uniform, Dense, and Sparse strategies, with Dense outperforming the others for both models. Specifically for Mixtral-8$\times$7B, Dense strategy achieves a 9.6\% reduction in Wikitext2 perplexity, whereas the other two strategies yield only around a 2\% improvement compared to the HQQ baseline. From the perspective of sparse layer's data distribution, we fix the rank of dense layer to 512 and compare the strategies of Uniform, Kurtosis, and Frequency. The Kurtosis strategy shows good performance improvement on both models since it captures weights with more outliers and with larger quantization error as analyzed in Fig.~\ref{fig:kurtosis-vs-error}. Frequency is a fairly good strategy, which brings more improvement to those models with unbalanced expert frequency, e.g. DeepSeek-MoE. These results have demonstrated the importance and opportunities of designing mixtures of adaptive low-rank compensators for a variety of MoE models.

\paragraph{Extra benefits from quantizing the low-rank compensators?} \textbf{Yes.} We compare the INT8 and INT3 compensators by evaluating Wikitext2 perplexity on Mixtral-8$\times$7B across a range of rank settings, as shown in \tref{tab:int3_int8}. Compared to INT8 compensators, INT3 only uses 37.5\% of memory resulting in just a 0.2\% increase in perplexity. Although there are occasional instances where the INT3 compensator causes a notable error surge in individual weights, as measured in Frobenius norm, the overall performance impact remains minimal. Overall, INT3 compensators achieve memory savings with negligible performance loss, aligning well with our motivation and methodology.

\begin{table}[!ht]
  \centering
  \small
  \caption{INT8/INT3 low-rank compensator results on Wikitext2 PPL for Mixtral-8$\times$7B.}
    \begin{tabular}{lcccc}
    \toprule
    Rank  & \multicolumn{2}{c}{\name Compensator Memory} & \multicolumn{2}{c}{Wikitext2 PPL$\downarrow$ } \\
    \midrule
          & INT8  & INT3  & INT8  & INT3 \\
    16    & 296 MB   & 106 MB  & 4.5014 & 4.5084 \\
    32    & 525 MB   & 212 MB  & 4.4682 & 4.4786 \\
    64    & 983 MB   & 424 MB  & 4.4054 & 4.4174 \\
    \bottomrule
    \end{tabular}%
  \label{tab:int3_int8}%
\end{table}%

\begin{figure}[!ht]
    \centering
    \includegraphics[width=0.8\linewidth]{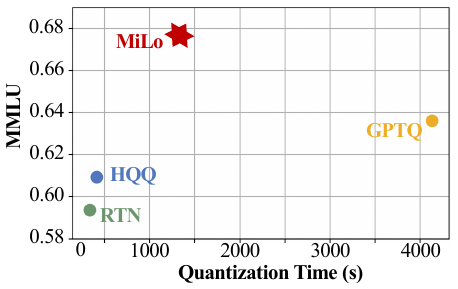}
    \caption{Quantization time vs. MMLU accuracy.}
    \label{fig:enter-label}
\end{figure}

\paragraph{Does \name add high compression overhead?} The INT3 quantization time versus the MMLU for Mixtral-8$\times$7B is plotted in Fig.~\ref{fig:enter-label}, with \name iterations set to 20. As a calibration-free method, our method gives 3$\times$ speedup compared to GPTQ while delivering the best accuracy. Although \name is slower than the other two calibration-free methods, HQQ and RTN, it remains within an acceptable timeframe.

\begin{figure*}[!ht]
    \centering
    \includegraphics[width=\linewidth]{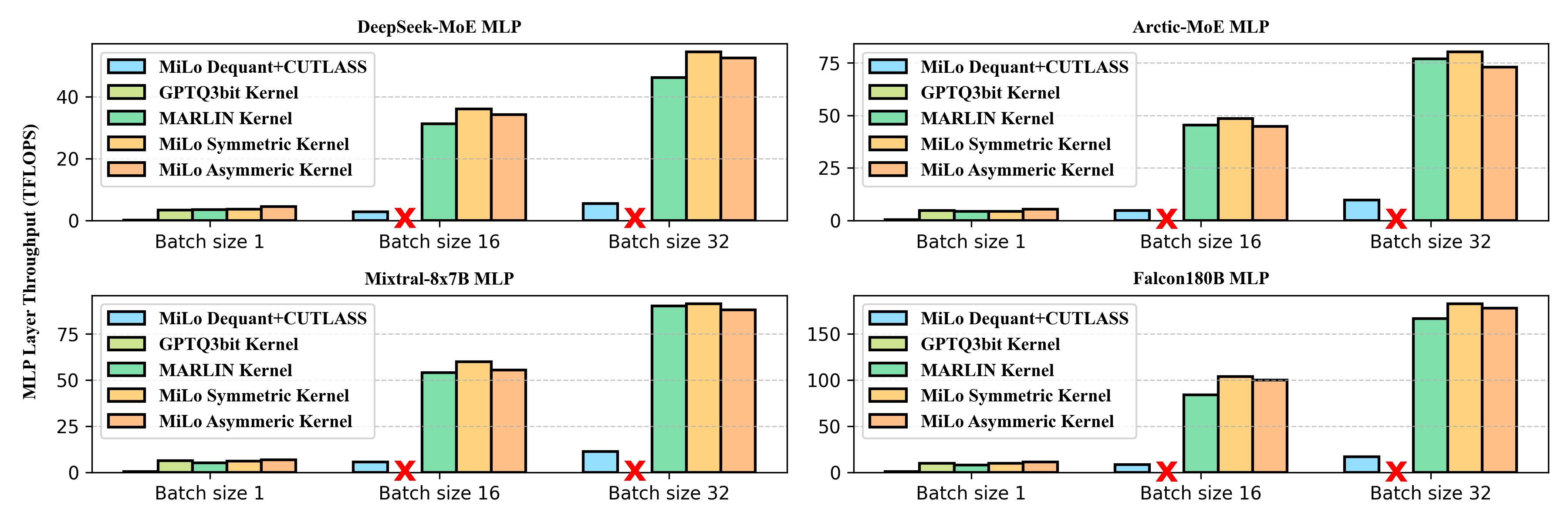}
    \caption{GeMM TFLOPS results on different model MLP layer.}
    \label{fig:GEMM}
\end{figure*}
\subsection{System Performance Results}

We evaluate our system performance through three components: end-to-end latency benchmarking on Mixtral-8$\times$7B, mixed-precision GeMM throughput (TFLOPS) analysis on the MLP layers of various models, and an ablation study to assess the impact of individual optimizations. All the system performance experiments are performed on an NVIDIA A100 GPU with 40GB memory.

\subsubsection{End-to-end Performance}
We perform MiLo algorithm on the Mixtral-8$\times$7B model and compare the end-to-end latency using three different backends, and we also bring the un-quantized model as a reference, with results shown in Table~\ref{tab:latency}. We consider the following backend settings: (1) PyTorch, which runs the un-quantized model, aiming at showing the effectiveness and the necessity of quantization and optimized backends. (2) GPTQ3bit Backend, which uses the kernel introduced in \cite{frantar2022gptq}. It realizes INT3 $\times$ FP16 GeMV kernel, which only supports quantized inference with batch size 1 and asymmetric per-channel quantization setting. (3) MARLIN Backend, which uses the MARLIN kernel introduced in \cite{marlin}. It provides a highly optimized INT4 symmetric per-channel quantization. (4) MiLo Backend, which uses the INT3 × FP16 GeMM kernel introduced in \sref{subsec:int3-kernel}. \name backend supports both symmetric and asymmetric quantization for batched inference and fine-grained quantization. In this comparison, we choose asymmetric quantization and a group size of 64 for \name backend, which provides better model accuracy but also adds additional computational overhead.

The PyTorch baseline runs out of memory because the Mixtral-8$\times$7B model takes $\sim$90GB memory, which exceeds the VRAM of an A100 GPU. GPTQ3bit Backend shows similar behavior with \name Backend at batch size of 1, but fails to support larger batch size settings. Compared with MARLIN Backend, \name Backend delivers a 1.2$\times$ speedup on batch size 1 and 1.26$\times$ speedup when batch size larger than 1, thanks to its INT3 quantization and the efficient asymmetric quantization support. 

The relative performance speedup (compared with MARLIN Backend) is better than the corresponding speedup results in GeMM throughput test reported in \sref{subsec:GeMM}, because \name quantization algorithm is asymmetric while MARLIN kernel does not inherently support this setting. When integrating MARLIN kernel for this experiment, we need to handle the zero-point calculations separately, which brings extra computation overhead. By fusing the asymmetric de-quantization operation and GeMM into a single kernel, \name kernel reduces additional traffic to GPU global memory and brings extra speedups.

\begin{table}[!ht]
\centering
\caption{End-to-end latency for Mixtral 8$\times$7B.}
\begin{tabular}{lccc}
\toprule
Backend / Batch size & 1 & 16 & 32 \\
\midrule
PyTorch       & OOM   & OOM   & OOM \\
GPTQ3bit Backend      & 0.102 & --    & --    \\
MARLIN Backend          & 0.123 & 0.141 & 0.145 \\
\name Backend & \textbf{0.102} & \textbf{0.112} & \textbf{0.113} \\
\bottomrule
\end{tabular}
\label{tab:latency}
\end{table}

\subsubsection{Mixed-Precision GeMM Performance}
\label{subsec:GeMM}
Fig.~\ref{fig:GEMM} shows the TFLOPS achieved by different mixed-precision GeMM solutions for the MLP layers of various models. The figure compares the following configurations: (1) MiLo Dequant + CUTLASS: It uses two operators. For the de-quantization, it uses MiLo Dequant introduced in \sref{subsec:int3-kernel}. For GeMM, it uses CUTLASS to do the calculation. The two operators are not fused. We use symmetric quantization with a group size of 64. (2) GPTQ3bit Kernel: It fuses asymmetric per-channel de-quantization and INT3 $\times$ FP16 GeMV, which is introduced in \cite{frantar2022gptq}. (3) MARLIN Kernel: It is the implementation in \cite{marlin} with fused symmetric de-quantization of a group size 128 and INT4 $\times$ FP16 GeMM. (4) MiLo Kernel: Results of both symmetric and asymmetric kernels described in \sref{subsec:int3-kernel} with a group size of 64 are reported. The detailed shape of the GeMM for this experiment can be found in Appendix~\ref{sec:gemm-shape}. When the batch size is 1, both \name Symmetric Kernel and the GPTQ3bit Kernel achieve the highest throughput. It is because GeMM here is highly memory-bound, and both solutions utilize 3-bit weights for data transfer, reducing memory overhead. For the batch size 16, we observe that \name Symmetric Kernel outperforms MARLIN Kernel, by 16\%, 7\%, 12\%, 24\% on MLPs of DeepSeek-MoE, Arctic-MoE, Mixtral-8$\times$7B, and Falcon180B, respectively. 
As the batch size increases to 32, the problem becomes compute-bound. Yet our kernel still demonstrates the highest throughput, with 17\% higher than the second best kernel on the DeepSeek-MoE MLP, due to the reduction in global synchronization overhead.

\subsubsection{Ablation Study}
We perform an ablation study on the MLP layers of five models using \name Asymmetric Kernel, to show the benefit brought by the optimization introduced in \sref{subsec:int3-kernel} including MiLo Dequant, Asynchronized global weight load, and MoE-specific tile shape tuning. The results are shown in Fig.~\ref{fig:kernel ablation}, where Baseline represents the \name Asymmetric Kernel. The experiment uses 3-bit quantization with a group size of 64, and the batch size is 16. The MLP sizes increase from left to right. We conclude that:

\begin{figure}[!ht]
    \centering
    \includegraphics[width=1\linewidth]{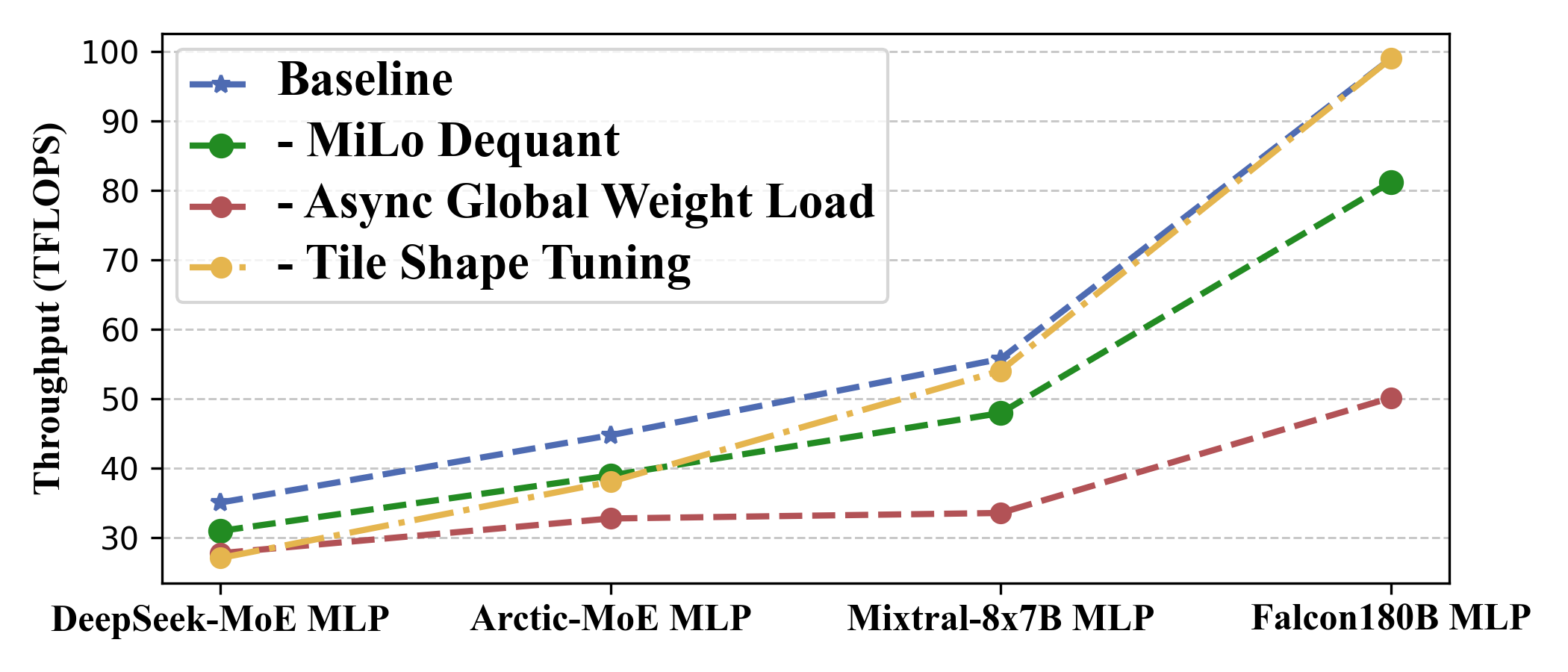}
    \caption{Ablation study of the proposed techniques.}
    \label{fig:kernel ablation}
\end{figure}

(1) Asynchronized global weight load proves to be the most critical component across all models, with its removal resulting in the largest performance degradation, because it overlaps weight loading with computation, significantly reducing pipeline stalls and maximizing GPU utilization.

(2) MiLo Dequant becomes increasingly important as the MLP size grows. Once Asynchronized global weight load is applied to overlap global memory transfers with computation, the remaining performance bottleneck shifts to the compute phase. This bottleneck becomes more pronounced in larger models, where MiLo Dequant effectively reduces the associated overhead.

(3) MoE-specific tile shape tuning has a significant impact when working with smaller matrices, such as those found in DeepSeek-MoE MLPs. However, its effect diminishes as the matrix size increases. This observation is consistent with our expectation that, for larger matrices, the relative cost of reduction operations is less significant compared to the dominant compute workload.

\section{Conclusion}

We present \name, a novel method that significantly improves the inference efficiency of MoEs, with negligible accuracy loss, using calibration-free quantization and mixture of low-rank compensators. We develop hardware-friendly W3A16 GeMM kernels for compressed MoE models, which delivers real latency reduction. Areas for future exploration include combining \name with other MoE compression techniques, such as pruning and distillation.

\section*{Acknowledgement}
We sincerely appreciate the insightful feedback from the anonymous reviewers. We thank Niranjan Uma Naresh for helpful discussions. This research was supported by the National Science Foundation (NSF) under Grant No. 2441601. The work utilized the Delta system at the National Center for Supercomputing Applications (NCSA) through allocation CIS240055 from the ACCESS program. ACCESS is an advanced computing and data resource program funded by the U.S. NSF under the Office of Advanced Cyberinfrastructure awards \#2138259, \#2138286, \#2138307, \#2137603, and \#2138296. The Delta advanced computing resource is a collaborative effort between the University of Illinois Urbana-Champaign and NCSA, supported by the NSF (award OAC 2005572) and the State of Illinois. This work utilized the Illinois Campus Cluster and NCSA NFI Hydro cluster, both supported by the University of Illinois Urbana-Champaign and the University of Illinois System.

\bibliography{reference}
\bibliographystyle{mlsys2025}

\appendix
\section{Trade-off Between Rank and Performance Gains}

\begin{figure}[!ht]
    \centering
    \includegraphics[width=0.75\linewidth]{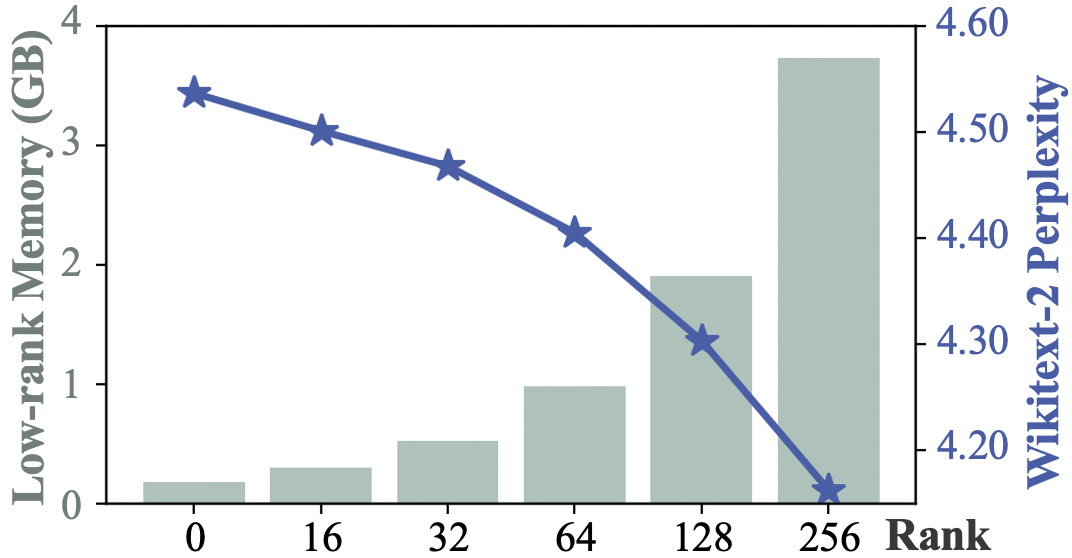}
    \caption{Additional memory consumption from using low rank compensators vs. perplexity varying the rank.
    }
    \label{fig:mixtral_rank}
\end{figure}

\fref{fig:mixtral_rank} illustrates the rank-accuracy relationship by comparing memory consumption and Wikitext2 perplexity as rank increases, emphasizing the trade-off between memory overhead and performance gains. 

\begin{table*}[htbp]
\centering
\small
\begin{tabular}{llcccccc}
\toprule
\textbf{Model} & \textbf{Rank Strategy} & \textbf{Wikitext2 PPL$\downarrow$} & \textbf{HellaSwag$\uparrow$} & \textbf{Lambada$\uparrow$} & \textbf{PIQA$\uparrow$} & \textbf{MMLU$\uparrow$} & \textbf{TriQA$\uparrow$} \\
\midrule
\multicolumn{8}{c}{\textit{Model Strategy, Memory constraint = 200MB}} \\
\midrule
\multirow{3}{*}{Mixtral-8x7B} 
  & Uniform-28     & 4.5262 & 79.63 & 74.01 & 80.63 & 61.58 & 72.15 \\
  & Dense-512      & 4.1683 & 81.00 & 72.34 & 80.84 & 65.75 & 74.71 \\
  & Sparse-32      & 4.5968 & 78.45 & 71.66 & 80.03 & 59.87 & 70.89 \\
\midrule
\multirow{3}{*}{DeepSeek-MoE} 
  & Uniform-22     & 6.9243 & 72.19 & 70.77 & 78.56 & 37.76 & 55.83 \\
  & Dense-512      & 6.4743 & 74.08 & 73.60 & 78.40 & 40.22 & 58.35 \\
  & Sparse-24      & 6.9770 & 71.51 & 66.56 & 78.12 & 35.93 & 54.55 \\
\midrule
\multicolumn{8}{c}{\textit{Sparse Layer Strategy (With Dense Layer rank = 512)}} \\
\midrule
\multirow{3}{*}{Mixtral-8x7B} 
  & Uniform-32     & 4.1645 & 81.13 & 72.07 & 81.17 & 66.67 & 73.95 \\
  & Kurtosis-32    & 4.1044 & 81.71 & 74.50 & 81.22 & 67.98 & 76.21 \\
  & Frequency-32   & 4.1698 & 80.95 & 77.21 & 80.84 & 66.47 & 75.28 \\
\midrule
\multirow{3}{*}{DeepSeek-MoE} 
  & Uniform-16     & 6.4633 & 73.90 & 72.92 & 78.56 & 40.45 & 58.41 \\
  & Kurtosis-16    & 6.3030 & 74.36 & 72.09 & 78.29 & 41.07 & 60.14 \\
  & Frequency-16   & 6.4570 & 73.61 & 69.55 & 78.89 & 38.22 & 58.84 \\
\bottomrule
\end{tabular}
\caption{Performance comparison of Mixtral-8x7B and DeepSeek-MoE across different rank strategies.}
\label{tab:rank_strategy_eval_more}
\end{table*}

\section{Implementation}
\label{appendix:implementation}
We use \texttt{Pytorch} in version 2.4.1+cu121 and \texttt{Transformers} in version 4.44.0 to implement our algorithm. The quantization is performed using functions from HQQ library, and low rank compensation is realized using function \texttt{torch.svd\_lowrank}, which approximates the largest singular values. The algorithm is implemented as described in Algo.~\ref{alg:QC}, and we use the early stop at 20 to terminate the iteration. The following experiments and evaluations are performed using \textit{lm-evaluation-harness}\footnote{\href{https://github.com/EleutherAI/lm-evaluation-harness}{https://github.com/EleutherAI/lm-evaluation-harness}}. We conduct experiments using a single NVIDIA A100 GPU with 40GB of memory.

\section{Shape of GeMM in Throughput Tests}
\label{sec:gemm-shape}

Below we list the shapes of the FFN layer matrices for different models used in our GeMM throughput experiments:

\begin{table}[!ht]
\centering

\caption{Matrices' shape in different model's FFN layer.}
\begin{tabular}{l c c}
\toprule
  & DeepSeek-MoE & Arctic-MoE \\
  \midrule
  w1 & (2048, 11008) & (7168, 4864) \\
  w2 & (11008, 2048) & (4864, 7168) \\
  w3 & (2048, 11008) & (7168, 4864) \\
  \toprule
  & Mixtral-8x7B & Falcon180B \\
  \midrule
  w1 & (4096, 14336) & (14848, 14848 $\times$ 5) \\
  w2 & (14336, 4096) & (14848 $\times$ 5, 14848) \\
  w3 & (4096, 14336) & -- \\
\bottomrule
\end{tabular}
\end{table}

\section{Correctness Test}

To ensure the correctness of our kernel, we conducted a comprehensive series of tests. These included \textbf{Functional Correctness Tests} to verify the kernel's basic operations, \textbf{Error Handling Tests} to evaluate its robustness against invalid or unexpected inputs, and \textbf{Boundary Conditions Tests }to assess its behavior at the extremes of operational parameters. The correctness criterion was defined as achieving a relative error of less than 0.005 across all tests, which were conducted using 5 different random seeds. The results demonstrated that our kernel passed all the tests successfully.

\textbf{Functional Correctness Tests}  
We evaluated functional correctness using real-world matrices. Specifically, in \texttt{test\_mixtral\_shape()}, we tested 4 different matrix shapes from the Mixtral8x7B model with batch sizes ranging from 1 to 1024. Similarly, in \texttt{test\_llama\_shape()}, we tested 16 different matrix shapes from the Llama2 model, again with batch sizes ranging from 1 to 1024. These tests confirmed that the kernel produced correct outputs across all configurations.

\textbf{Error Handling Tests}  
We verified the kernel's ability to handle errors under three specific conditions:  \\
1. The group size must be set to 64.  \\
2. The shape of the weight matrix \((k, n)\) must be a multiple of the tile shape \((64, 256)\), \((128, 128)\) or \((256, 64)\). \\  
3. The tile shape configuration must be restricted to \((64, 256)\),\((128, 128)\) or \((256, 64)\).  \\
These tests ensured the kernel could detect and appropriately respond to invalid configurations.

\textbf{Boundary Conditions Tests}  
Boundary conditions were tested on two dimensions: the batch size and the reduction dimension (the input dimension of the weight matrix).  \\
1. Batch Size Dimension: We focused on scenarios where the batch size is not a multiple of 16, as tensor cores perform 16x8x16 matrix multiplications. In such cases, padding is required to ensure compatibility when the batch size is not divisible by 16.  \\
2. Reduction Dimension: We examined cases where the reduction dimension is not a multiple of 4 * tile\_shape[0]. This is because we group 4 tiles into one pipeline calculated by the threadblock, and in these situations, the matrix handled by a threadblock during one pipeline stage terminates early.  

These tests confirmed that the kernel performs correctly and efficiently, even under edge cases and challenging scenarios.

\section{Rank Strategy Evaluations}
\tref{tab:rank_strategy_eval_more} presents a complementary evaluation of rank strategies, using Wikitext2 perplexity, zero-shot and few-shots evaluations. All the evaluations support the previous conclusion that dense layers deserve a higher rank compared with sparse layers, and the Kurtosis value is a good index to identify the sparse layers that deserve a higher rank.

\clearpage
\section{Artifact Appendix}
\subsection{Abstract}

This artifact description provides a comprehensive workflow for MiLo, which covers how to run the algorithm, rank strategy generation, and evaluation on publicly available benchmarks, along with the quantized kernel settings and evaluation.

We provide instructions on how to obtain, build, and run the software, as well as the necessary steps to reproduce the results presented in the paper. Additionally, the experiment scripts are editable to accommodate further implementations and verifications.

\subsection{Artifact check-list (meta-information)}

{\small
\begin{itemize}
  \item {\bf Algorithm: } The MiLo algorithm, which employs an iterative optimization to optimize quantized MoE models with a mixture of low-rank compensators.
  \item {\bf Program: }Python, CUDA
  \item {\bf Dataset: } For the evaluation, we include Wikitext2, PIQA, HellaSwag, Lambada, MMLU, and TriQA, all of which are publicly available through Huggingface.
  \item {\bf Hardware: } See \ref{hardware-dependency}.
  \item {\bf Metrics: } The metrics for algorithm include perplexity, accuracy, exact match, memory consumption, and execution time. The metrics for the backend kernel include TFLOPS and execution time.
  \item {\bf Output: } Quantization time and quantized model memory; Wikitext2 perplexity; zero-shot and few-shots benchmark accuracy.
  \item {\bf Experiments: }See \ref{sec:experiement-workflow}.
  \item {\bf How much disk space required (approximately)?: } 100GB
  \item {\bf How much time is needed to prepare workflow (approximately)?: } 15 min
  \item {\bf How much time is needed to complete experiments (approximately)?: } Basic experiments (quantization, perplexity, zero-shot tasks) take about 2 hours; Running all the experiments (including few-shot tasks) requires about 12 hours.
  \item {\bf Publicly available?: } Yes
  \item {\bf Code licenses (if publicly available)?: } MIT license
\end{itemize}}

\subsection{Description}

\subsubsection{Code Access}
The MiLo algorithm, benchmarks, and scripts are available at Github: 
\sloppy
\href{https://github.com/Supercomputing-System-AI-Lab/MiLo}{\texttt{Supercomputing-System-AI-Lab/MiLo}}
\fussy

\subsubsection{Hardware dependencies}
\label{hardware-dependency}
The MiLo algorithm should be able to execute on Nvidia GPUs with sufficient GPU memory (e.g., 40GB). The MiLO backend kernel is currently only compatible with the NVIDIA Ampere architecture (e.g., NVIDIA A100). For the algorithm, we recommend testing on an NVIDIA A100 GPU with 40GB/80GB memory. 

\subsubsection{Software dependencies}
The software is performed using Python 3.10, and CUDA version 12.4.0. The dependent Python packages can be found in the \texttt{requirements.txt} file. 

\subsubsection{Data sets}
The evaluation datasets include Wikitext2, HellaSwag, Lambada, PIQA, MMLU and TriQA, all of which are publicly available and can be downloaded from Huggingface.
\subsection{Installation}
First, please access the code by
\begin{verbatim}
$ git clone --branch MiLo-beta https://gith
ub.com/Supercomputing-System-AI-Lab/MiLo.git 
\end{verbatim}

To better reproduce and avoid capability issues, we recommend using Python 3.10 and CUDA version 12.4.0. 

We provide the scripts for the recommended environment setup. Please follow the instructions to create the Conda environment and install the MiLo package \& kernels.

\begin{verbatim}
$ conda create -n milo python==3.10
$ conda activate milo
$ bash conda_env_setup.sh
\end{verbatim}
And for the kernel setup, please run the following script:
\begin{verbatim}
$ bash kernel_setup.sh
\end{verbatim}
\subsection{Experiment workflow}
\label{sec:experiement-workflow}
\paragraph{MiLo quantization algorithm experiments.}
MiLo provides bash scripts to reproduce the results from the paper. The main results for the quantization and evaluation of Mixtral-s1 in Table \ref{tab:main_result} can be reproduced by executing the bash scripts provided as:
\begin{verbatim}
$ cd MiLo
$ bash examples/Mixtral_s1.sh <YOUR_DIR>
\end{verbatim}
Please change \texttt{<YOUR\_DIR>} to your local directory to save the model. This script includes MiLo quantization, evaluation on Wikitext2 perplexity, and zero-shot evaluation. These experiments take around 2 hours. The similar scripts are provided for Mixtral-s2 and DeepSeek-s1 and DeepSeek-s2 in the examples folder. 

The few-shot evaluations can be executed using a separate script as:
\begin{verbatim}
$ bash examples/MiLo_fewshots_eval.sh <YOUR_
DIR> <MODEL_NAME>
\end{verbatim}
Please change \texttt{<YOUR\_DIR>} to your quantized model directory and \texttt{<MODEL\_NAME>} to 
"DeepSeek" or "Mixtral". Please note that this might take 10 hours to run on an NVIDIA A100 40GB GPU, and we separate the evaluation for the convenience of testing.
\newline


\paragraph{MiLo INT3 kernel experiments.}
MiLo provides bash scripts to reproduce the results presented in the paper. The kernel GeMM throughput results shown in Figure \ref{fig:GEMM} can be obtained by running the following bash script:
\begin{verbatim}
$ cd MiLo
$ bash examples/kernel_GeMM_performance.sh
\end{verbatim}
Similarly, the kernel end-to-end latency results reported in Table 6 can be reproduced using the following script:
\begin{verbatim}
$ cd MiLo
$ bash examples/kernel_end2end_latency.sh
\end{verbatim}
Due to recent code modifications, some additional overhead has been introduced, leading to slight deviations from the results presented in the paper. To ensure the validity of our work, we also provide results from the state-of-the-art INT4 kernel, Marlin, for comparison. Please note that this might take an hour on an A100 GPU since we need to install MARLIN and quantize the Mixtral-8x7B model to INT4.
\subsection{Evaluation and Expected Results}
\paragraph{MiLo quantization algorithm experiments.}
The evaluation scripts will print the quantization time and quantized model memory to the terminal output. 
The evaluation results will be saved to \texttt{<YOUR\_DIR>/eval\_result.json}.

\paragraph{MiLo INT3 kernel experiments.}
We provided four scripts for the four kernel experiments in correspondence, as listed below. The expected results are included within the scripts.  
\begin{enumerate}
    \item 
GeMM correctness results under different settings. The correctness is checked by default. An assertion error will be triggered on incorrect output.
    \item 
GeMM throughput results.  
    \item 
Customized GeMM throughput results. This test supports the group size 64 quantization setting and \texttt{tile\_shape} {(64, 256), (128, 128) and (256, 64)}.  
    \item 
End-to-end first token latency. The results correspond to Table \ref{tab:latency} in the main text of the paper.  
\end{enumerate}

\subsection{Experiment customization}
\paragraph{MiLo quantization algorithm experiments.}
By editing the bash script, you can experiment with customized quantization configurations to examine the quantization algorithm.

For example, you can modify the launching command as below to quantize the Mixtral-7x8b model with a uniform rank of 32:
\begin{verbatim}
python utils/MiLo_quant_main.py 
    --base_dir <YOUR_DIR> 
    --model_id Mixtral 
    --dense_rank 32
    --sparse_rank 32
\end{verbatim}

\paragraph{MiLo INT3 kernel experiments.}
Here we show how to test the MiLo INT3 kernel with varying matrix configurations. 

For example, you can launch the example script as below to collect the GeMM results using batch size 16, weight output dimension 7168, weight input dimension 2048, tile shape (128, 128):
\begin{verbatim}
$ bash examples/kernel_custom_GeMM.sh 
    --batch_size 16 
    --weight_output_dimension 7168 
    --weight_input_dimension 2048
    --tile_shape 128,128 
\end{verbatim}


\subsection{Methodology}

Submission, reviewing and badging methodology:

\begin{itemize}
  \item \url{http://cTuning.org/ae/submission-20190109.html}
  \item \url{http://cTuning.org/ae/reviewing-20190109.html}
  \item \url{https://www.acm.org/publications/policies/artifact-review-badging}
\end{itemize}

\end{document}